# Complex picking via entanglement of granular mechanical metamaterials

*Ashkan Rezanejad, Mostafa Mousa, Matthew Howard, Antonio Elia Forte\**

A. Rezanejad, M. Mousa, M. Howard, A. E. Forte

Department of Engineering, King's College London, London, WC2R 2LS, UK
E-mail: antonio.forte@kcl.ac.uk

**Abstract:** When objects are packed in a cluster, physical interactions are unavoidable. Such interactions emerge because of the objects' geometric features; some of these features promote entanglement, while others create repulsion. When entanglement occurs, the cluster exhibits a global, complex behaviour, which arises from the stochastic interactions between objects. We hereby refer to such a cluster as an "*entangled granular metamaterial*". We investigate the geometrical features of the objects which make up the cluster, henceforth referred to as grains, that maximise entanglement. We hypothesise that a cluster composed from grains with high propensity to tangle, will also show propensity to interact with a second cluster of tangled objects. To demonstrate this, we use the entangled granular metamaterials to perform complex robotic picking tasks, where conventional grippers struggle. We employ an electromagnet to attract the metamaterial (ferromagnetic) and drop it onto a second cluster of objects (targets, non-ferromagnetic). When the electromagnet is re-activated, the entanglement ensures that both the metamaterial and the targets are picked, with varying degrees of physical engagement that strongly depend on geometric features. Interestingly, although the metamaterial's structural arrangement is random, it creates repeatable and consistent interactions with a second tangled media, enabling robust picking of the latter.

# 1. Introduction

Mechanical metamaterials possess unique properties due to their specific geometric architecture. Some of these attributes are rarely found in ordinary materials [1–4]. Such engineered structures are usually continuous and characterised by a repeating pattern. However, disordered metamaterials, which have non-repeating patterns, are often found in nature and can provide enhanced properties [5] including superior toughness showed in animal bones [6–8]. Additionally, when the metamaterial's cells are disconnected and interact as independent "grains", interesting behaviours may emerge from the cluster, including collective mechanical properties in the form of adaptive [9], programmable [10], and jamming [11] granular media. Furthermore, different grain shapes, such as beams, rods [12,13] or ellipsoids [14,15], further influence the formation of the cluster. Structures which form bird's nests are a natural example of clusters made from beam grains [16]. These special types of disordered mechanical metamaterials have been recently labelled as "granular metamaterials" [17], the grains of which show high degree of interaction in confined spaces, resulting in the emergence of a collective behaviour (e.g., granular viscosity).

Here we aim to identify geometrical features that cause entanglement, driving the grains to behave collectively. Entanglement usually occurs in a cluster of interacting objects with peculiar shapes [18–20]. For instance, spiky objects are inherently prone to tangle, but longer spikes reinforce separation [21]. In particular, we hypothesise that a cluster formed by grains with high propensity to tangle will also show propensity to interact with a second cluster of tangled objects. To demonstrate this, we develop a new robotic picking method. Specifically, we employ an electromagnet to attract the granular metamaterial (ferromagnetic) and drop it onto a second cluster of objects (targets, non-ferromagnetic). The grains penetrate, inter-lock and secure the targets in part or in whole. When the electromagnet is re-activated, the entanglement between the metamaterial and the targets ensures that both are picked. Subsequentially, they can be released at a second location by deactivating the electromagnet. Additionally, we developed manufacturing strategies to create decomposable grains, which leads to entanglement deactivation and separation of targets and metamaterial.

Recent strategies have exploited entanglement to achieve functionality in robotic picking and grasp rigid and (mainly) singular objects [22,23]. However, there have been no attempts to control the picking of entangled media accurately and consistently. Indeed, such media are extremely challenging to pick consistently with conventional fingered robotic grippers [24–28], (e.g., parallel, 3-finger or anthropomorphic hands), suction devices, scoops, electromagnets, or universal grippers (i.e., granular-jamming type), [29].

Differently, we here unlock a new strategy to pick highly deformable and disordered interacting objects, such as wires, cotton threads, woollen yarns, herbs, and netting, with high consistency. Moreover, by reducing the control parameters to a minimum and harnessing the flexibility of metamaterial design, our method opens new pathways for robotic gripping and manipulation, simplifying the control system and embedding functionality into the hardware.

## 2. Results and Discussion

### 2.1. Effect of geometry on granular entanglement

A granular media is a collection of grains that can have uniform or arbitrary properties. Depending on these properties, the grains might show several levels of interaction. We hypothesise that the grain's geometrical features play a major role in these interactions, resulting in higher propensity to entanglement. Therefore, we start by examining the entanglement behaviour emerging from clusters of grains with different shapes.

For simplicity, we choose nine generic grain shapes, or types (see **Figure 1A**). Each grain is created from a combination of straight or curved 12mm segments. For details about the grains' fabrication, refer to Section S1, Supporting Information. The simplest grain is made of only one base segment (type i). Other types are created by adding spikes, which are perpendicular segments to the base segment (or to its projection in the case of curved base segments). Therefore, we generate grains with one-spike (type ii, iii and iv), two-spikes (type v, vi and vii) and four-spikes (type viii and ix).

#### 2.1.1. Packing fraction

In our first experiment, we measure the packing fraction [21] of the different grain types. The experimental set up consists of a hollow cylinder (diameter 30 mm, height 100 mm) filled with 4 g of one grain type at a time (corresponding to 100 segments). Firstly, we manually shake the set up for 10 seconds to enhance interactions and compact the granular media (see Figure 1B), and we measure the height of the granular structure inside the cylinder (i.e., $h_0$). Then we calculate the packing fraction as the ratio of the volume occupied by the grains ($V_g$ – approximately the same for all grain types as 100 segments of each is used) to the volume of the structure inside the cylinder ($V_c$ – differs for each grain type, as this is based on the height of their granular structure, $h_0$).

### 2.1.2. Integrity test

To demonstrate that grains with different geometries have different degrees of entanglement, we run a second experiment to measure the structural integrity of different granular media [30] as a proof-of-concept. The experimental set-up follows the same procedure as the previous experiment. However, after recording the initial height of the granular structure ($h_0$), the cylinder is removed with a gentle, vertical movement, leaving behind a structure of clustered grains (see Figure 1C), the height of which depends on the degree of entanglement relative to the specific grain type. The new height of such structures is recorded and the change in height is defined as $\Delta h$. The structural integrity is measured as the ratio of standing height compared to the initial height, mathematically expressed as: $(h_0 - \Delta h) / h_0$.

It is noticeable that different grain types show different self-supporting capabilities, offering resistance against gravity, which is generated from the internal interactions among single grains. In particular, some types collapse immediately due to the lack of entanglement (type i) or excessive number of spikes (types viii and ix), which results in repulsion rather than interlocking. Others show a higher degree of entanglement, which triggers self-supporting internal forces, opposing gravity. Quantitative results are reported in Figure 1F, depicted as mean values and standard deviation for a set of 10 repetitions. Interestingly, type v grains form the structure with the highest degree of integrity of 95% (followed by type iv and vii, which show 89% and 88% structure integrity, respectively) indicating that such shapes lead to minimum repulsion between grains, while ensuring maximum entanglement.

### 2.1.3. Robotic gripping test

To translate this concept into robotics and demonstrate how entanglement affects robotic picking we run a third experiment in which we pick a cluster made of a single grain type with a robotic parallel gripper (see Figure 1D). This type of gripper uses force closure to pick objects with accessible surfaces. Since in granular media the grains are randomly clustered and do not form a clear contact surface with which the parallel gripper can engage, there is a large variance across the 10 picking iterations (see Figure 1G). However, the difference in the mean picked amount between type v grains and the other types is evident (more than 10% higher than type iv – second highest picked grain type, and 25% higher than type vii – third highest picked grain type). This is due to the high entanglement capability of grain type v, which results in picking a cluster that extends outside the grasp of the parallel gripper.

### 2.1.4. Entangled granular metamaterials

It is important to notice that the U-shaped grains (type iv, v and vii) have a collective, complex behaviour that emerges when interacting in a disordered cluster, which is likely to be caused by their shape. Such behaviour differs substantially from that shown by other grain types, where global characteristics fail to emerge. For this reason, we name clusters formed by these types of grains "entangled granular metamaterials".

For simplicity, we continue the analysis that follows with employing exclusively type v grains, as they show the highest degree of entanglement, under the hypothesis that this grain type will also create interactions with entangled targets (i.e., the items we want to pick). From now on, we will refer to them simply as "grains".

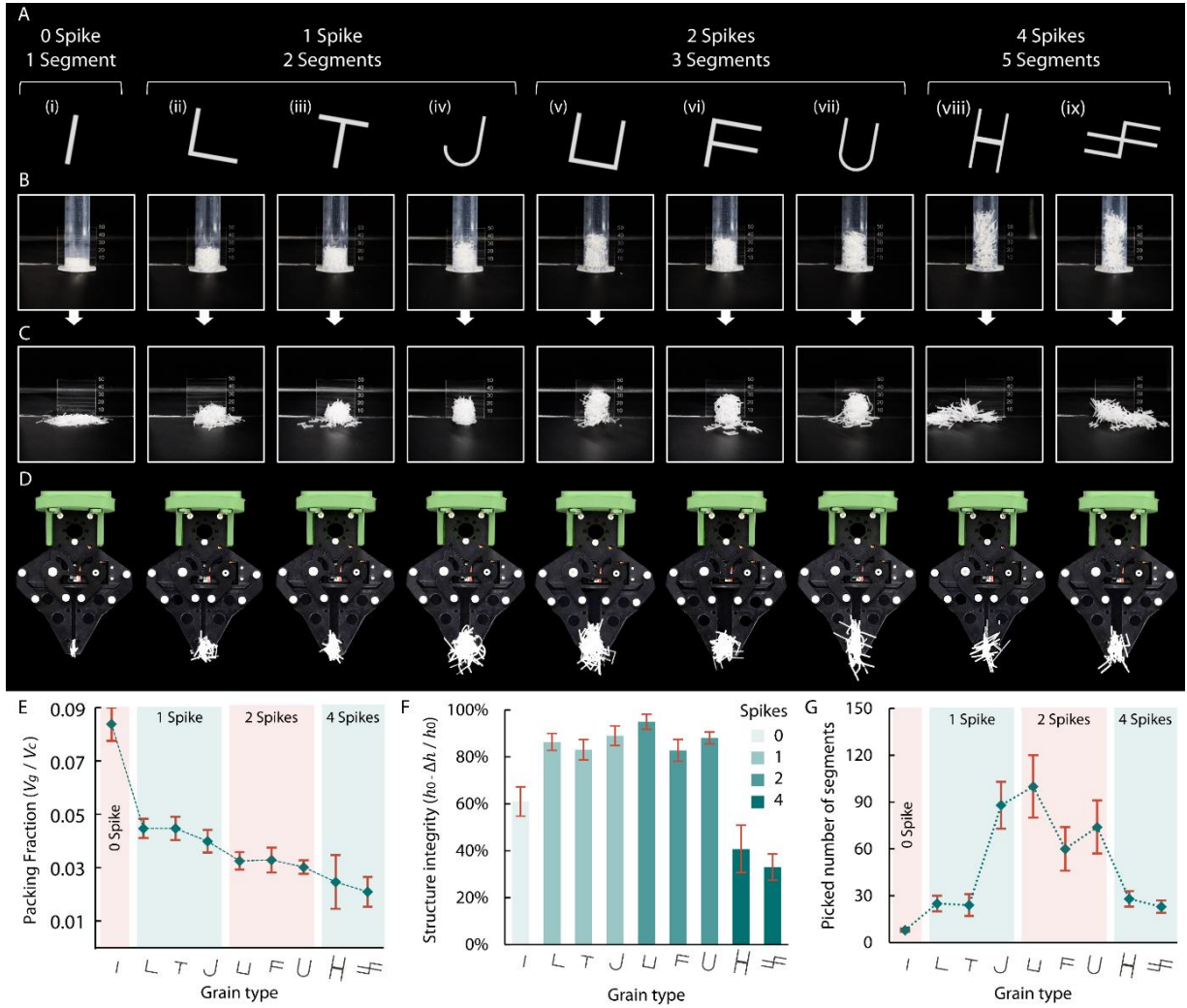

**Figure 1.** Collective behaviour according to geometric features. **(A)** Nine grain shapes with different number of segments and spikes, where a single bar represents one segment (i). Here we call "spike" a segment that extends perpendicularly from the base segment and has one free end. The base segment can be straight (ii, iii, v, vi, viii and ix) or curved (iv and vii). **(B)** Real-life images of 100 segments (4 grams) of each grain type packed in a cylindrical tube. **(C)** The self-supporting structure of the grains after removal of the cylindrical tube. **(D)** Real-life images of picked grains with a robotic parallel gripper. **(E)** Packing fraction of each grain type calculated via the ratio of occupied volume by the grain ($V_g$) to the cylindrical volume with the same height as the granular structure height ($V_c$). **(F)** Structural integrity of each grain type measured by the ratio of the structure height after removal of the cylindrical tube ($h_0 - \Delta h$) to the initial structure height ($h_0$). **(G)** Picked number of segments from clusters of 150 segments (6 grams) per grain type by a robotic parallel gripper.

## 2.2. Parametric study of geometrical features of entangled target cells

We now aim to use the grains' ability to entangle when interacting in a disordered cluster to create a momentary structure that can pick clusters of entangled objects (i.e., targets).

To investigate the picking performance, we designed an experimental set-up comprising a robotic arm, an electromagnet mounted on the robotic end-effector, ferromagnetic grains (stainless steel type v grains), a cluster of target cells packed in a container, and a laboratory scale to measure the number of picked target cells with a resolution of 0.01 g (refer to Section S2, Supporting Information for further details).

The experimental procedure begins with the robotic arm approaching a container filled with 100 ferromagnetic grains and picking these up by activating an electromagnet mounted on its end-effector (Figure 2A (1,2) and Figure 2B). The robotic arm then moves its end-effector above the targets and deactivates the electromagnet, resulting in a sudden drop of all the picked grains onto the cluster of entangled target cells (Figure 2A (3)). The grains penetrate the cluster and create entanglement (Figure 2C). Afterwards, the electromagnet is re-activated to attract all the grains once again. This results in the picking of both the grains and the target cells, as they are entangled together (Figure 2D). Lastly, the picked entangled media are moved to the desired position and dropped into another container by deactivating the electromagnet (Figure 2A (4)).

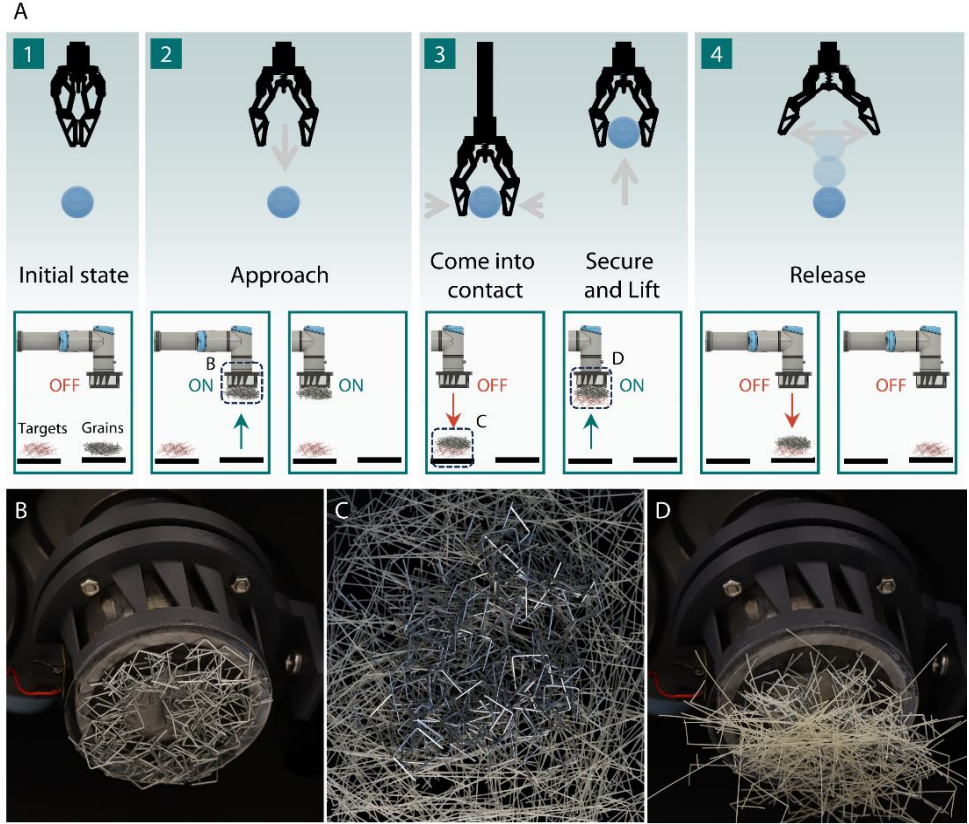

**Figure 2.** Grasping process and entangled picking. **(A)** A comparison of the step-by-step picking process of a conventional parallel gripper (top) to our grasping approach (bottom). **(B)** Real-life image of electromagnetic end-effector picking grains. **(C)** Real-life image of entanglement between the grains and the target. **(D)** Real-life image of entangled picking of the target by the grains.

Next, we shift our attention to the target's geometry and investigate which features affect entanglement. We hypothesise that thickness ($\tau$), length ($\lambda$) and the number of spikes ($\sigma$) in picking targets are the main geometric parameters that govern entanglement (see Figure 3A). We therefore vary these parameters in our fourth experiment to reveal any effect they might have on picking performance. In particular, we fabricate targets using all possible combinations of the parameters $\tau$ = [0.2, 0.4, 1] mm, $\lambda$ = [12, 60, 120] mm and $\sigma$ = [0, 1, 2]. This results in 27 types of target cells (see Figure 3B for examples of target cells with same thickness, $\tau$ = 1 mm). All the target cells have a uniform cross-sectional thickness along their length (see Figure S7), and a fixed length of 12 mm for all spikes. The cells were designed using CAD software (Autodesk Fusion 360) and fabricated by laser cutting manufacturing methods (see Section S3, Supporting Information for details).

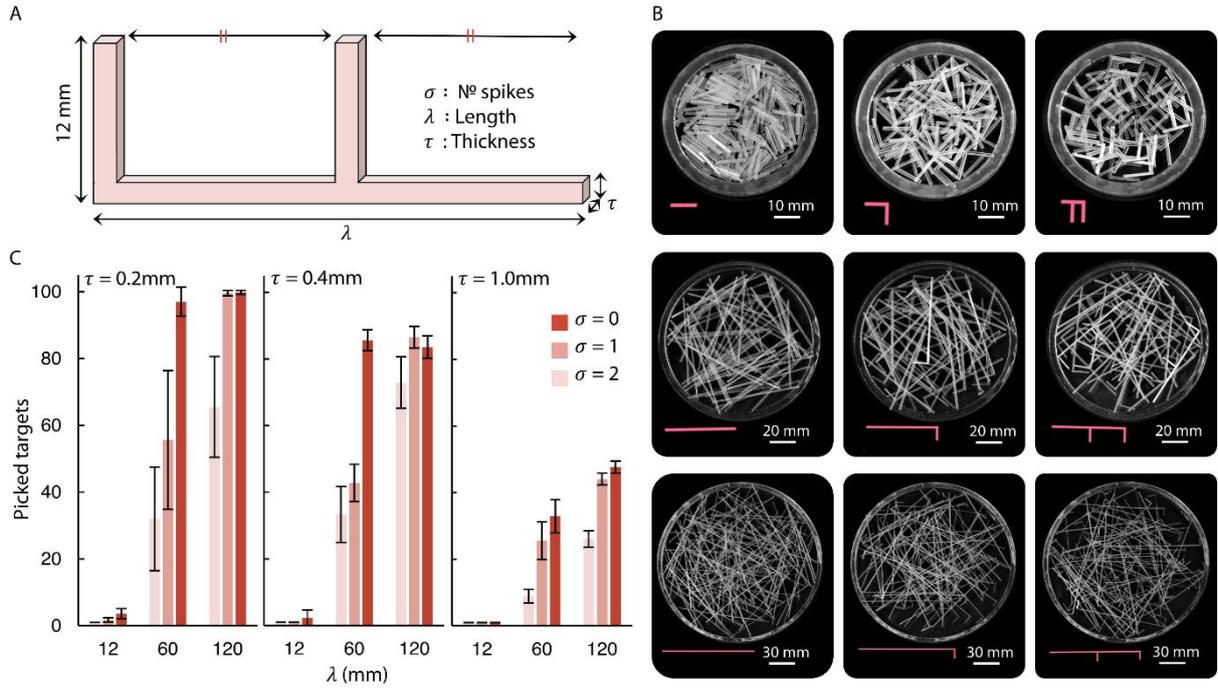

**Figure 3.** Geometrical parametric study for engineered target cells. **(A)** Target cell's geometrical parameters. **(B)** Real-life images of the target cell clusters. The scale reference is included at the bottom-right corner of each image. **(C)** Picked average number of targets and standard deviation across the investigated parameters.

To promote packing of the targets in a repeatable way, we choose a curved container (i.e., plastic bowl, see Figure S5), from which to pick the target cells, with decreasing internal diameter (from top to bottom). Because of this, the targets tend to slide to the bottom of the bowl due to gravitational effects, ensuring that they are all within the reach of the electromagnet. The bottom of the container has the lowest internal diameter of 80 mm – which is the same as the diameter of the electromagnetic robotic end-effector allowing full accessibility. The container's diameter gradually increases towards the top, with a maximum diameter of 200 mm (66% larger than the longest target cell). Furthermore, we choose a relatively powerful electromagnet (with maximum pull force of 2000 N at zero distance) to ensure that every single ferromagnetic grain is picked in this experiment.

When comparing the picking performance of various target cells, we measure the number of units picked, out of 100 units in the container. The picking results suggest that longer, thinner, and more spiky objects are significantly more prone to tangle, where length is the parameter with the greatest impact (see Figure 3C). According to the dominance analysis for four-dimensional data [31], length has the highest importance (74.4%) in picking performance, while thickness and number of spikes have smaller impact (16.4% and 9.1%, respectively). Refer to Section S4, Supporting Information for the details of the statistical analysis.

Additionally, we run a supplementary experiment to investigate the effect of bending stiffness on picking performance, which is reported in Section S5, Supporting Information. Based on the results from statistical tests and for simplicity, we hereby neglect the effect of bending stiffness in the following investigation.

## 2.3. Probabilistic model to predict entangled picking based on the geometrical parameters

Having observed the impact of geometric parameters on entangled picking using our grasping approach, we develop a probabilistic model which can predict the picked amount based on targets' thickness, length and spikiness. Specifically, the model assumes a linear relationship between the cells' length ($\lambda$), and the picked amount ($\hat{y}$), modulated by a logistic function of the cells' thickness ($\phi(\tau)$). Mathematically it is expressed as:

$$\hat{y} = \omega_1 \, \phi(\tau) \, \lambda + \omega_2 \tag{1}$$

We train our model by optimising the linear regression coefficients ($\omega_1$ and $\omega_2$) and the logistic function features (refer to Section S6, Supporting Information for further details) using the dataset from the previous experiment on target's geometry. Separate models are trained for (a) non-spiky (0-spike) and (b) spiky targets (1-spike and 2-spikes) since these exhibit substantially different behaviours. Despite limited data, and the highly random and disordered nature of the entangled media, this simple model achieves low errors (NMSE < 0.12) on unseen picking data with high confidence (see Figure S9) indicating our ability to control picking for previously unseen targets.

## 2.4. Picking real-life targets and validating the entangled picking model

To test our grasping approach performance and accuracy of our prediction model, we choose a variety of real-life targets that entangle when in a cluster. All our real-life targets are usually longer than 100 mm. For this reason, we fix the length across all tested targets to 120 mm as this is the highest length in our control study. This enables our model to make a more accurate prediction. We experiment with five real-life targets (see Figure 4A (i-v)), three of which are non-spiky (cotton threads (i), cables (ii) and knitting wool (iii)) and two that are spiky (fresh parsley (iv) and mint (v)). Furthermore, to demonstrate an application beyond the tested approach, we test our picking accuracy for an entirely different type of target which has a mesh form (walnuts in a net bag, Figure 4A (vi)).

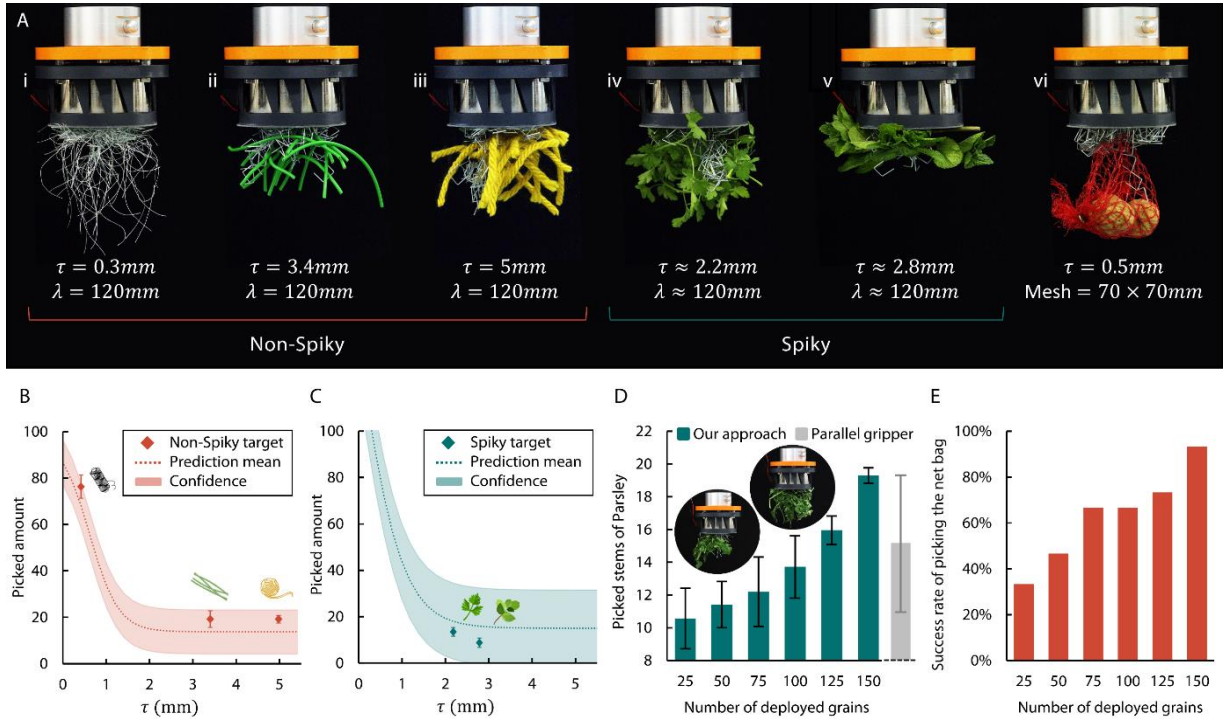

**Figure 4.** Picking real-life targets. (**A**) Images of picking non-spiky, spiky and meshed real-life targets (varying in thickness) cotton thread (i), cables (ii), knitting wool (iii), fresh parsley (iv), fresh mint (v) and walnuts in a net bag (vi). (**B**) The average picked number of "non-spiky" real-life targets and the prediction model (confidence represents the standard deviation of the prediction) with respect to the thickness. (**C**) The average picked number of "spiky" real-life targets and the prediction model with respect to the thickness. (**D**) The mean and standard deviation of picked stems of fresh parsley (by varying the number of deployed grains from 25 to 150) in comparison to the picked mean and standard deviation obtained with a parallel gripper. (**E**) The success rate chart of picking net bags in respect to the number of deployed grains.

The result reported in Figure 4B and 4C show that the picking of spiky and non-spiky real-life targets lie within the prediction region of our probabilistic model (with a close proximity to the prediction mean). Moreover, the small variation in the picked amount of targets across 10 iterations indicates high reliability and control over picking.

To compare our grasping approach with a conventional one, we tested picking performance of a commercial robotic parallel gripper (Actobotics Parallel Gripper Kit A) when picking entangled real-life targets (refer to Section S7, Supporting Information). The result shows that our approach has consistently lower standard deviation in picking compared to a parallel gripper (around two times (96%) lower on average when 100 grains are deployed) and offers controllability over the picked amount (see Figure S7 (C)).

Additionally, we study the impact of the number of deployed grains on the picked targets using our approach. We choose fresh parsley as a target example to be picked by grains. We then vary the number of grains from 25 to 150 and measure the amount of picked parsley across 10 iterations (see Figure 4D). The result shows that varying the number of grains allows control of the picked amount – with deployment of a larger number of grains corresponding to higher number of picked targets, which is a feature that other robotic grippers usually lack. In fact, as shown in Figure 4D, the parallel gripper can only pick one average number targets, with very high standard deviation of ~27%. This is 2 times and 11 times the standard deviation we achieve with our approach when deploying 25 grains and 150 grains, respectively.

Lastly, we study the success rate of picking meshed objects (walnuts in a net bag). The chart reported in Figure 4E demonstrates a maximum picking success rate of 95% when 150 grains are deployed (in 20 iterations of picking). In order to run a comparison with a parallel gripper, a more complex set up would be needed, including cameras and vision-based algorithms. Differently, our approach is straightforward, with minimal controls.

**2.5. Design and testing of decomposable grains**

Here we demonstrate two types of decomposable grains to attain complete release of the picked target by dismantling or melting the grains (refer to Section S8, Supporting Information for details of fabrication).

### 2.5.1. Ferromagnetic dismantlable grains

The first type is constructed from a water-soluble casing and ferromagnetic metal bars. The casing is 3D printed using BVOH (Butenediol vinyl alcohol co-polymer, Ultrafuse BVOH, BASF) due to its exceptionally quick disintegration in water, as compared to other commercially available 3D printing filaments such as PVA (Polyvinyl Alcohol). The metal bars are firmly fitted and secured in the casing (see Figure 5A) and the picking experiment is run. The picked media is then dropped in water. When the constructed grains come in contact with water, the casing disintegrates, and the metal bars are released. Due to their shape (Figure 1) and density, the short metal bars are prone to untangle from the target and sink in the water (see Figure 5B and 5C).

### 2.5.2. Ferromagnetic ice grains

The second proposed design is constructed by mixing water with food-safe ferromagnetic powder (commercially known as "black iron oxide" with the chemical composition of $Fe_3O_4$, EU food number: E172, The Crafters Shop, UK) in a ratio of 2 to 1. The liquid mixture is poured into a moulding case and then placed inside a freezer to produce spiky ferromagnetic ice as our grains (see Figure 5D). An additional advantage is that this type of grain does not necessarily require an electromagnet, since the quick melting time causes the grains to disintegrate, resulting in the release of the picked targets. To demonstrate this, we replace the electromagnet with a neodymium magnet (RS components, with a pull force of ~400 N, see Figure 5E and 5F) for this last experiment. This reduces the set-up cost, minimises the control parameters, and allows for the iron oxide to be collected on the magnet and reused to create new grains.

The result reported in Figure 5G shows a comparison between the picked stems of fresh parsley by the two decomposable grain types and the previously used stainless steel grains (25 grains deployed). Both decomposable grain types pick around 7 stems of parsley which is only about 23% lower than the averaged picked amount obtained when using the stainless-steel grains.

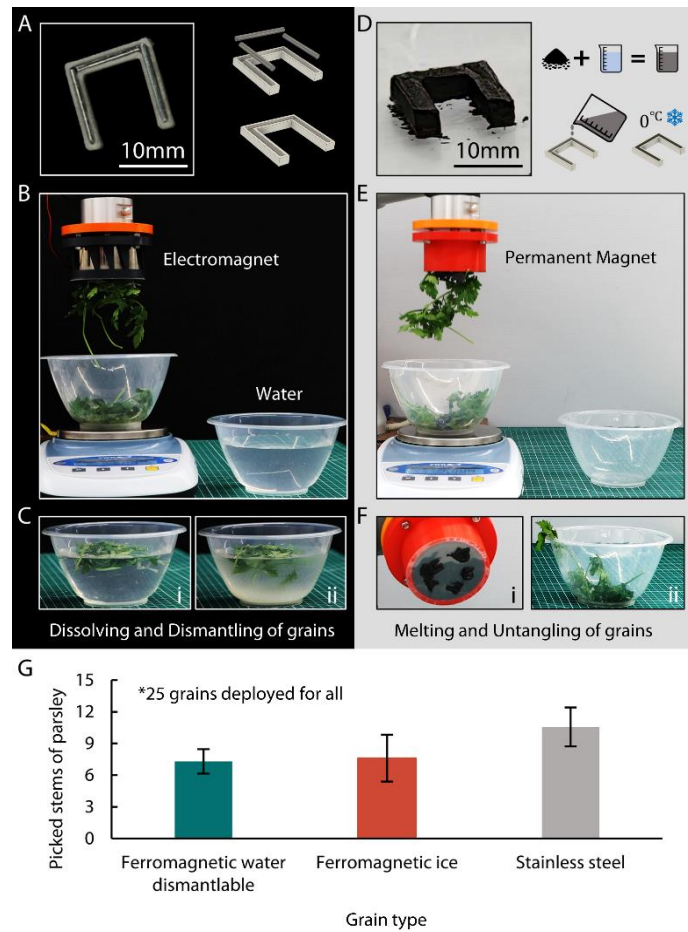

**Figure 5.** Fabrication and testing of decomposable Grains. **(A)** Ferromagnetic dismantlable grains constructed from stainless steel bars and 3D printed BVOH casing **(B)** Picking fresh parsley using an electromagnet and ferromagnetic dismantlable grains. **(C)** Targets and ferromagnetic dismantlable grains are dropped into water (left); the BVOH casing of the grains dissolves when in water resulting in dismantling and sinking of the metal bars (right). **(D)** Ferromagnetic ice grains constructed from water and ferromagnetic iron oxide particles. **(E)** Picking fresh parsley using a permanent magnet and ferromagnetic ice grains. **(F)** The remaining ferromagnetic powder collected on the magnet after the grains are melted (left); the targets released in the dropping zone (right).

## 3. Conclusion

Current grasping approaches [29] excel in picking individual and mainly rigid objects while struggling with consistent and accurate picking of multiple objects entangled in a cluster. The most commonly used gripping technique, "impactive" gripping (e.g., parallel grippers), strongly relies on the position and orientation of the tip of the jaws while picking. However, due to the disordered nature of entangled matter, such gripping systems struggle to find the optimal position and orientation, as they only have two or a few contact points to pick objects. Thus, impactive grippers cannot control the picked amount consistently [27,28], attaining either no picking or excessive picking. "Astrictive" grippers (e.g., vacuum suction grippers) require (i) high suction forces and/or (ii) a seal to be formed on the picked object (e.g., with a suction cup) to ensure sufficient force for lifting [32]. Due to these requirements, such grippers are not suitable to pick entangled targets, which do not offer smooth surfaces to create the sealing.

Other gripping methods such as "ingressive" (e.g., scooping grippers) can pick granular media [33] but struggle with entangled grains where they only achieve excessive picking with no consistency. Lastly, "contiguitive" grippers (e.g., chemical adhesion, freezing plate) can only pick objects with a clear surface and may cause permanent damage in the process, making them unsuitable for handling entangled targets such as fresh herbs.

In this work we have presented a novel approach for robotic manipulation that harnesses momentary structures. These are created by grains, the shape of which gives them the ability to tangle with each other and with other objects. Because of this emerging global behaviour, we have named this clustered media "entangled granular metamaterial". We have then shown how to harness this behaviour to pick a variety of real-life targets, by releasing a specific number of grains onto them. The grains first interact and interlock with the targets, and are then recalled by an electromagnet, which provides the magnetic force to pick up both grains and targets.

In order to achieve this, we designed a series of experiments to (i) unravel the grain shape with the highest degree of granular interaction, (ii) harness the tangling behaviour of such grains in robotic picking of entangled targets, (iii) investigate the effect of geometric features of such targets on picking performance, and lastly, (iv) develop a probabilistic model to predict the picked amount given such geometric features.

We considered the geometric features of targets (thickness, length and number of spikes) as the main parameters that affect entangled picking (see Figure 3C). We further demonstrated that it is possible to predict the picked amount only using the mentioned geometric parameters (see Figure 4B and 4C). We recognise that other material properties such as mass density, surface friction and bending stiffness, to some level, could affect entanglement. However, we hereby focused on investigating the effect of geometry, which we demonstrated to play a major role in triggering entanglement behaviour in granular media. Future studies could explore additional parameters (e.g. friction, material properties, etc.) to develop a more accurate predictive model for our picking approach.

Here, we have employed magnetic force as a stimulus to construct the entangled granular metamaterial from ferromagnetic grains. However, it is important to note that our gripping technique is general and can make use of several external stimuli (e.g. buoyancy, airflow, etc.) to achieve the picking task. This constitutes potential opportunity for further work, in which gripping prototypes actuated by other stimuli can be designed and tested.

Lastly, we presented ferromagnetic grains that can be decomposed, allowing complete separation from targets after picking. We designed two types of decomposable grains and demonstrated their performance against stainless steel grains (see Figure 5). They exhibit a slightly lower picked amount but a similarly low variance in picking. This further proves the potential of our picking approach in practical applications where accuracy is required. Nevertheless, there are some limitations with the current design of decomposable grains as they need (i) to be dropped in water in the case of ferromagnetic dismantlable grains and (ii) to be removed from the magnet during each picking iteration in the case of ferromagnetic ice grains.

Our gripping approach has significant potential for industrial applications, such as packaging lines for the food industry, which is falling behind from the rapid growth of robotic automation [34]. Despite the available technologies [35–37], there is a limited number of robotic solutions that tackle complex pickings — i.e., picking of entangled horticultural products which still largely rely on manual labour [38]. Furthermore, the gripping approach hereby presented could be employed for the handling of clusters of wires, tubes, pipes, or fibres. Indeed, we aim at customising our grasping approach (for example, varying grain shape and size) to offer a reliable robotic solution for these complex industrial tasks.

In conclusion, we identify the benefits of our gripping approach as having (i) reduced necessary control parameters, (ii) reconfigurability of entangled granular metamaterial to control the picked amount, and most importantly, (iii) the capability of picking complex and disordered clusters (entangled media) both consistently and accurately. These benefits unlock a new range of possibilities for robotic manipulation, enabling sophisticated tasks by harnessing disordered physical behaviours, which emerge from collective mechanical properties.

# 4. Methods

## 4.1. Design and fabrication of grain cells and target cells

We design 2D sketches of both grains and target cells using a commercial CAD software (Autodesk Fusion 360). We laser-cut (Universal Laser System) 1 mm thick acrylic sheets (Polymethyl methacrylate, Hobarts) to fabricate both grains (see Figure S2) and 1mm thick target cells. However, we use 0.25 mm and 0.1 mm mylar sheets (biaxially oriented polyethylene terephthalate, RS components), for fabricating the remaining target cells (0.4 and 0.2 mm thick). This was done for two reasons: (i) acrylic sheets with thickness below 1 mm were not commercially available and (ii) mylar's friction coefficient (~0.5) and density (1.4 g/cc), [39] are close to acrylic (0.5 and 1.2 g/cc, respectively), [40], making it a good alternative for our experiments. Through multiple trial-and-error tests, we determine the optimal laser cutting settings to achieve uniform cross-sectional thickness in our target cells. This was done to compensate for the sub-millimetre melting that occurs while laser cutting thin mylar sheets. For further details refer to Section S3, Supporting Information.

## 4.2. Electromagnetic robotic gripper and control

The electromagnet mounted on the robotic end-effector (UR3 robotic arm) is commercially available (RS PRO, access control door magnet) with 2000 N maximum pull force, 12 V DC and maximum generated current of 1 A. Both the connector to robot's end effector and the casing for the electromagnet are designed in CAD software (Autodesk Fusion 360) then 3D printed using FDM (fused deposition modelling) with an Ultimaker-S5 3D printer (using PLA (polylactic acid) filament), see Figure S3 for details.

Additionally, we use a laboratory scale (VWR® Science Precision Balances) to measure the picked amount with resolution of 0.01 g and a sensitivity of ±0.03 g. We also use an Arduino Mega board, a 12V DC power supplier and a 12V to 5V voltage relay to control the electromagnetic gripper. The overall control of the system including the robotic arm is programmed using ROS (robotic operating system) and Python scripting, see Figure S4 for more details including the flow chart and connection diagram.

The amount of target cells (Figure 3C), silicon tubes (Figure S8) and real-life targets (Figure 4B-D, Figure 5G and Figure S10) picked are reported as the number of units picked (for example number of target cells in the geometric study or number of stems of parsley in real-life targets). To this end, the mass of each unit cell is measured as a reference. Then, the recorded mass of picked amount by the laboratory scale is divided by a single unit cell's mass to find the number of units picked accordingly.

### 4.3. Statistical analysis and probabilistic prediction model

All the results reported in this paper are the mean and standard deviation of 10 repetitions of experiment. This is done to capture the stochastic nature of entangled clusters. To better understand the results, we have performed a suitable statistical analysis (four-dimensional-data dominance analysis) to show the relative importance of three geometric parameters on picking using our gripping approach. This method measures the interactional dominance, which represents the parameter's incremental impact in the presence of all other predictors. This is obtained by subtracting the R-squared value of a model with all other parameters from the R-squared value of the complete model. Furthermore, it finds the Relative Importance of each predictor according to its interactional dominance [31]. Refer to Section S4, Supporting Information for the analysis.

Moreover, we design two probabilistic prediction models (for non-spiky and spiky targets) which can approximately predict the picked amount given target's average thickness and length. As our geometric study is limited to a range of dimensions, our predictor performs best when the targets are within that range (12 to 120 mm length and 0.2 to 1 mm thickness). However, we show that the picked amount of thicker real-life targets (2.2, 2.8, 3.4 and 4 mm thick), which lie outside the range, is still well predicted. Refer to Section S6, Supporting Information for details regarding the mathematical derivation with demonstrative graphs of both models.

### 4.4. Fabrication of decomposable grains

#### 4.4.1. Ferromagnetic dismantlable grains

Firstly, we design a 3D model of dissolvable casing using CAD software (Autodesk Fusion 360). Then the part is printed using a direct extrusion FDM (fused deposition modelling) method with Prusa-MK4 3D printer. The water-soluble material used for 3D printing is BVOH (Butenediol vinyl alcohol co-polymer, BASF) filament. Afterwards, three 10 mm stainless steel bars are tightly fitted into the BVOH casings (see Figure S11).

### 4.4.2. Ferromagnetic ice grains

We start by fabricating a silicon moulding case for the ice grains. To do so, we 3D print a PLA mould for the silicon. Then, we pour fast curing flexible silicon (Eco-flex 00-50, Smooth-On) to cast the silicon moulding case. Thereafter, a well-mixed mixture (2 to 1 ratio) of water and black iron oxide ($Fe_3O_4$, The Crafter Shop, UK) is poured into the silicon moulding case and transferred to a freezer with temperature below 0 degrees to freeze the mixture (see Figure S12). The ferromagnetic ice grains are then manually extracted from the silicon mould. To avoid melting during extraction the grains are kept at subzero temperature before being employed in the robotic task.

**Supporting Information**

Supporting Information is available from the Wiley Online Library or from the author.

**Acknowledgements**

We would like to thank the technician team at King's College London: Charlotte Plamer, Samual Piper, Jon West and Alejandro Ramirez Mancebo who supported this research with their knowledge and time. Furthermore, we would like to thank Barakat Ibrahim Barakat and Yuxuan Wang who helped with fabricating target cells, videography and post-processing of images taken from the experiments. Lastly, the authors would like to acknowledge the financial support of UK research and innovation: UKRI grant EP/X525571/1 (A.E.F), UKRI grant MR/X035506/1 (A.E.F).

**Table of Contents (ToC)**

Current robotic grippers struggle when handling a cluster of entangled objects, which can be highly deformable and are also hard-to-model. Here we employ entangled granular metamaterial (ferromagnetic), to perform consistent robotic picking of any other entangled objects (non-ferromagnetic). This demonstrates how entanglement in granular media can exhibit a collective behaviour depending on the geometrical features of the grains.

**Title**: **Complex picking via entanglement of granular mechanical metamaterials**

**Authors:** Ashkan Rezanejad, Mostafa Mousa, Matthew Howard, Antonio Elia Forte*

**ToC Figure**

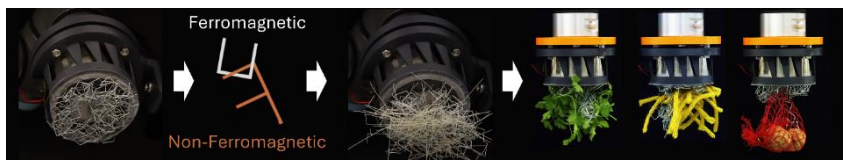

# Supporting Information

**Complex picking via entanglement of granular mechanical metamaterials**

*Ashkan Rezanejad, Mostafa Mousa, Matthew Howard, Antonio Elia Forte *

## Section S1. Fabrication of grains

To achieve a diverse range of grain attributes, we create three distinct grain subsets, each characterized by unique spike configurations (three core shapes of L-shaped, T-shaped and J-shaped). Each core shape is then used to branch out more designs and produce ultimately 26 grain designs, varying in number of spikes, orientation and configuration (see Fig. S1). For this study and for the sake of simplicity, we select nine grain designs and fabricate them with the laser cutting manufacturing method (Universal Laser System, cut from acrylic sheet, Polymethyl methacrylate, Hobarts, see Fig. S2).

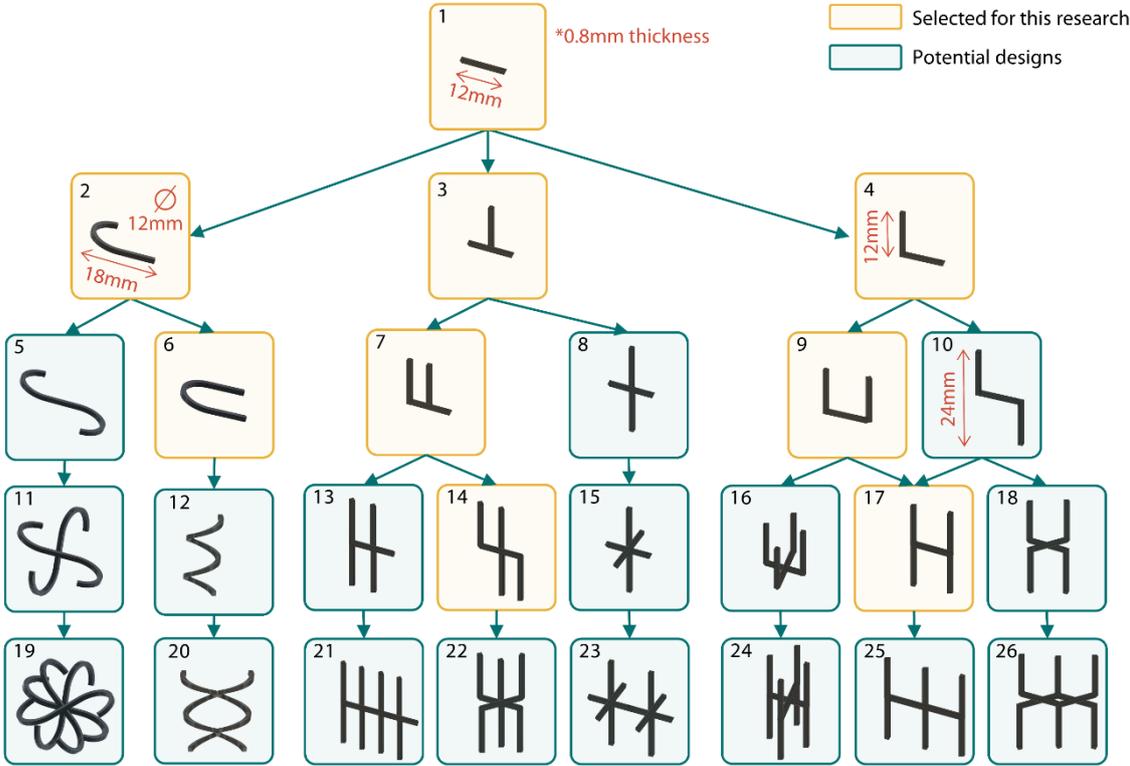

**Figure S1.** Illustration of families of grains. The yellow-highlighted grain types are the ones used in the integrity test and robotic gripping test reported in the result section.

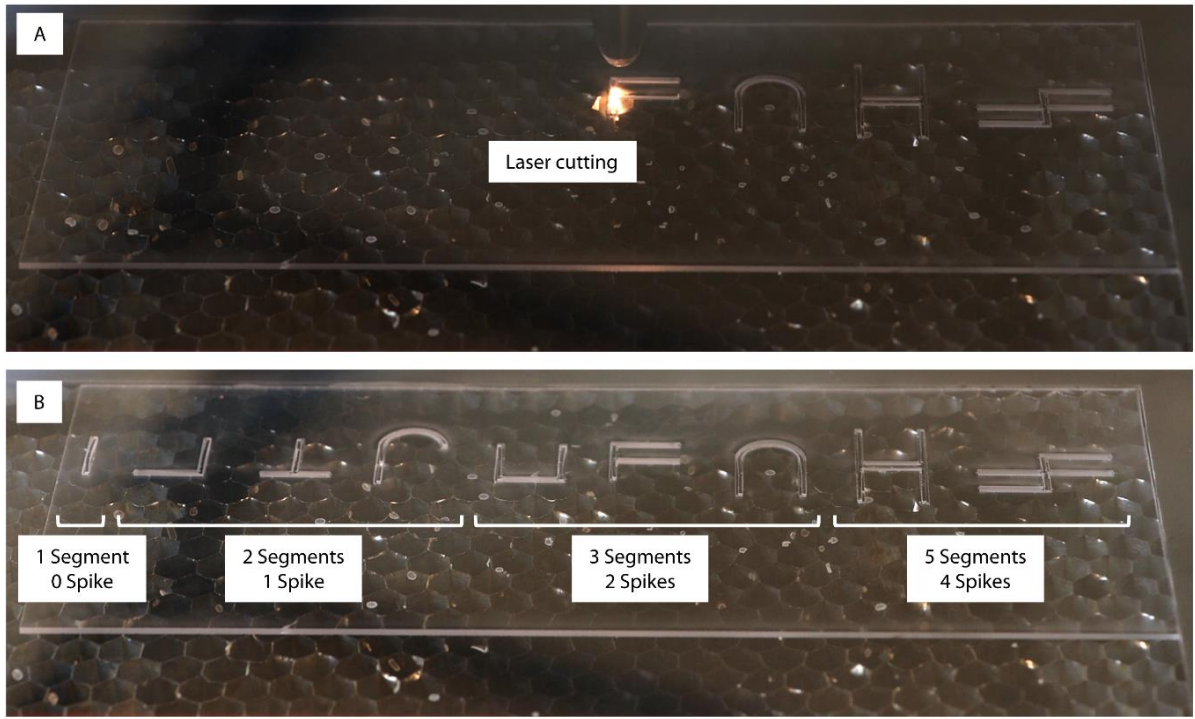

**Figure S2.** Fabrication of nine types of grains. All grains are fabricated by laser-cutting on a 1 mm thick acrylic sheet.

**Section S2. Experimental setup**

In our picking experiment, we choose an electromagnet (791-7558, access control door magnet, RS components) or in case of ferromagnetic ice grains (refer to supplement S8) a permanent magnet (811-2628, neodymium magnet, Eclipse, RS components) as a tool for manipulating the grains. We design (by Autodesk Fusion 360) and 3D print (by Ultimaker S5) casings for both the electromagnet and neodymium magnet. We then assemble and attach them to the end effector of UR3 (Universal Robot 3) robotic arm (see Fig. S3). Furthermore, control process (in form of a flow chart), connection diagram between the robot and the control unit, and components for the control unit are shown in Fig. S4. We also further report the step-by-step real-life images of the experimental procedure in Fig. S5.

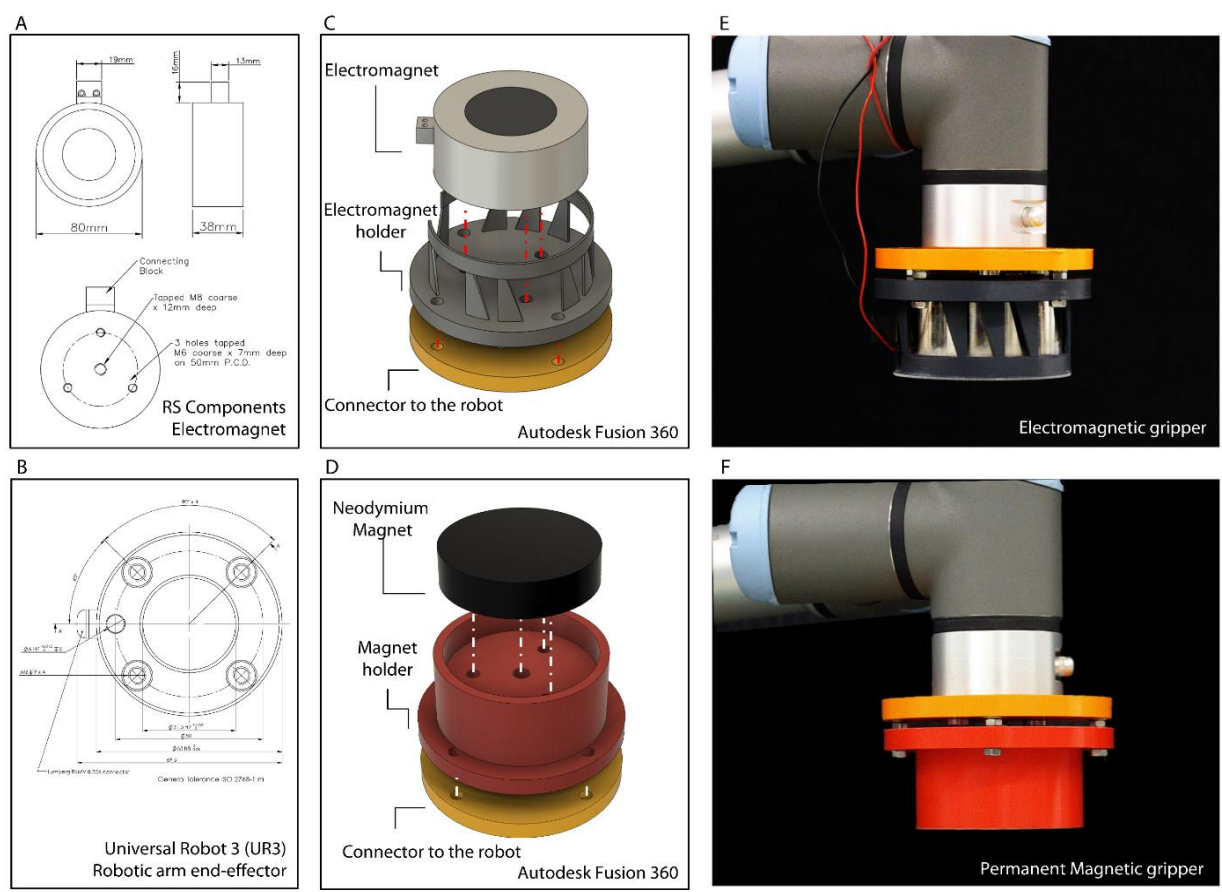

**Figure S3.** The assembly and design of magnetic robotic end-effector. (**A**) Dimensions of the electromagnet. (**B**) Dimensions of the robotic end-effector. (**C**) The assembly of the electromagnet into the holder and connector to the robotic end-effector. (**D**) The assembly of the Neodymium magnet into the holder and connector to the robotic end-effector. (**E**) The real-life image of the mounted electromagnetic gripper. (**F**) The real-life image of the mounted permanent magnetic gripper.

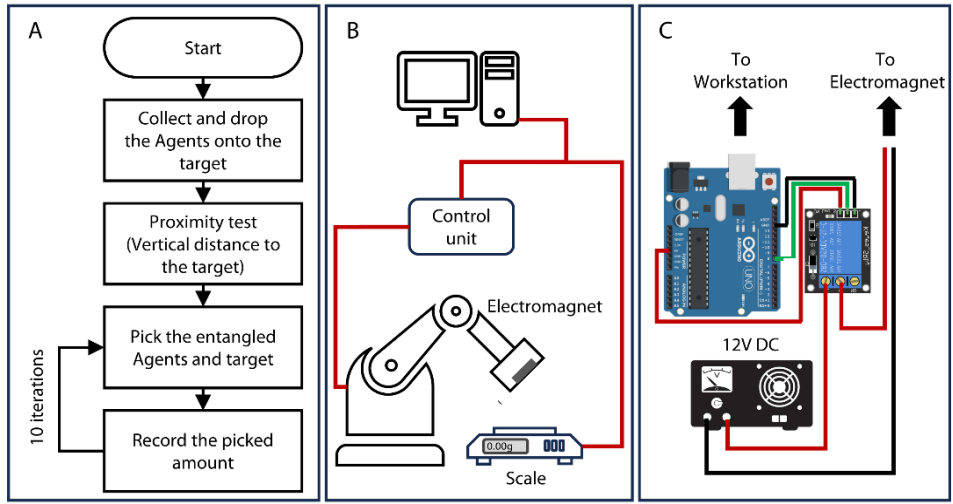

**Figure S4.** The control process. (**A**) The flow chart describes one set of experiment which includes 10 iterations of picking. After each iteration, the picked amount is dropped to be picked again for the next iteration. (**B**) The connection diagram of the experiment's main components; including an electromagnet, a laboratory scale, and a control unit. (**C**) The control unit's components; including an Arduino board connected to a 5 V voltage relay which controls the supply of 12 V to the electromagnetic gripper.

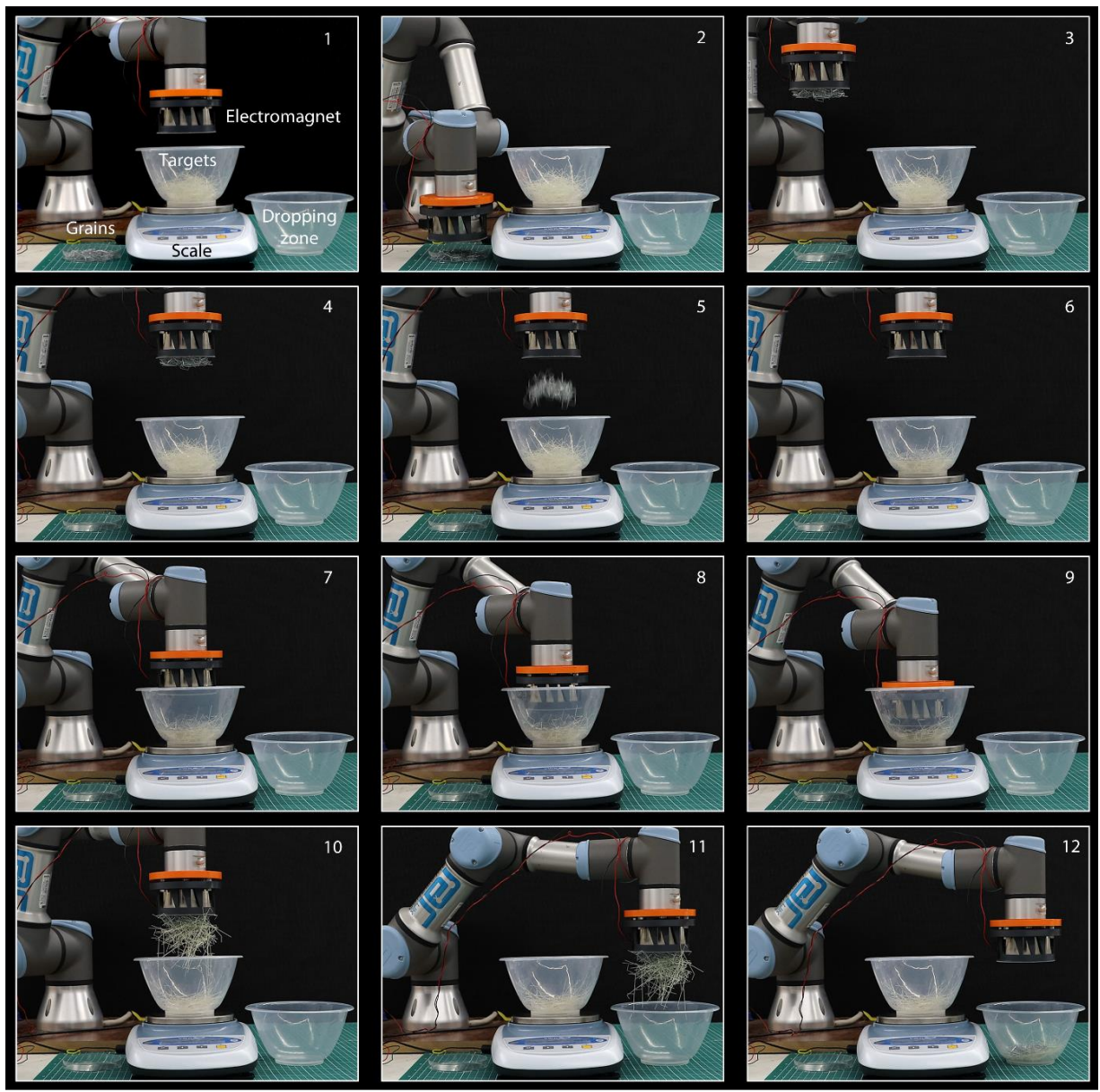

**Figure S5.** The experimental procedure. The real-life image of the experiment set-up including robotic arm (Universal Robot 3), a laboratory scale (VWR® Science Precision Balances), a container filled with grains, a plastic bowl filled with fabricated target cells and an empty plastic bowl as a dropping zone (1). The robotic arm approaches (the robot moves with a fixed velocity of 0.05 m/s in all steps) the grains and upon the activation of the electromagnet the grains are picked (2,3). The grains are transferred on top of the bowl of targets and upon the deactivation of the electromagnet are dropped into the targets (4-6). A proximity test is carried out to find the correct z coordinate (the vertical distance from the robotic end effector to the pile of mixed targets and grains) (7,8). The robot approaches the pile of mixed grains and targets and upon the activation of the electromagnet, a portion of the targets entangled to the ferromagnetic grains are picked (9,10). The picked pile is transferred to the dropping zone (empty bowl) and is dropped by deactivating the electromagnet (11,12).

**Section S3. Fabrication of target cells**

We report the real-life images of 27 fabricated target cells (varying in length, thickness and number of spikes) and their design sketch in Fig. S6.

We use laser cutting manufacturing method using Universal Laser System to fabricate the target cells. The laser cutting approach gives us control over two dimensions of cutting. Thus, we carefully select the thickness of the material sheets as it determines the third dimension of the target cell. We use 0.1 mm and 0.25 mm mylar sheets (Polyethylene terephthalate, RS components) as the low and medium levels of thickness, and 1 mm acrylic sheet (Polymethyl methacrylate, Hobarts) as the high level of thickness. We chose mylar sheets due to the lack of availability of acrylic sheets with a thickness of lower than 1 mm in the commercial market. Both acrylic and mylar have relatively similar material density and Young's modulus (as stated in Table S1), making mylar a suitable alternative. Furthermore, the bending stiffness (EI) is reported in Table S1, obtained by the product of Young's modulus (E) and the area moment of inertia (I) of its cross-section [41]. Mathematically, it can be expressed as: $EI = E \times I$.

**Table S1.** Material properties of target cells.

| Target cell | Material type | Thickness of the material sheet | Density* | Young's Modulus* (E) | Moment of Inertia (I) | Bending Stiffness (EI) |
|---|---|---|---|---|---|---|
| 0.2 mm | Mylar | 0.10 mm | 1.4 $\frac{g}{cm^3}$ | 2.8 − 3.5 GPa | 0.0008 mm$^4$ | ≈ 0.3 mN. m$^2$ |
| 0.4 mm | Mylar | 0.25 mm | 1.4 $\frac{g}{cm^3}$ | 2.8 − 3.5 GPa | 0.0012 mm$^4$ | ≈ 4.1 mN. m$^2$ |
| 1.0 mm | Acrylic | 1.00 mm | 1.2 $\frac{g}{cm^3}$ | 2.5 − 3.5 GPa | 0.1220 mm$^4$ | ≈ 366 mN. m$^2$ |

*Material property for Mylar [39] and Acrylic [40].

Furthermore, we report the program setting of the laser cutter for fabricating target cells in Table S2. The numbers reported are achieved after a set of trial and error aiming to cut target cells with uniform cross-sectional thickness (see Fig. S7 for microscopic images, Hitachi TM4000Plus). Due to melting and remaining residues of cut material during the process of laser cutting, it was necessary to try different setting parameters in order to achieve consistent and accurate cutting.

**Table S2.** Laser cutting setting properties according to the material thickness.

| Material | Desired thickness | Power | Pulses per inch | Speed | Z-axis |
|---|---|---|---|---|---|
| Mylar | 0.2 mm | 4 Watts | 1000 | 0.050 $^m/_s$ | 0.3 mm |
| Mylar | 0.4 mm | 24 Watts | 1000 | 0.038 $^m/_s$ | 0.5 mm |
| Acrylic | 1.0 mm | 42 Watts | 1000 | 0.015 $^m/_s$ | 1.2 mm |

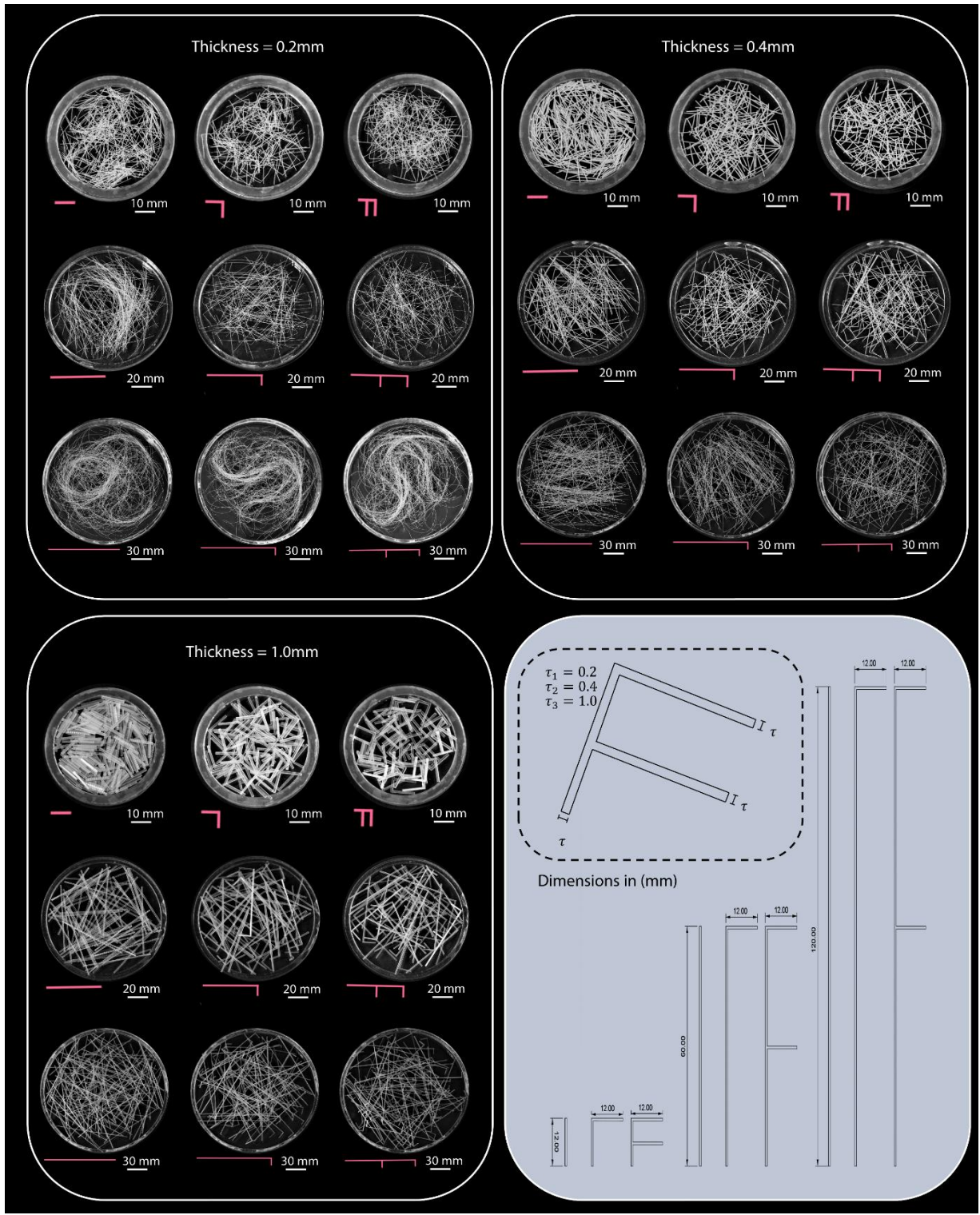

**Figure S6.** The real-life images of the fabricated target cells. Three values are assigned for three geometric parameters of: number of spikes (0, 1 and 2), length (12, 60 and 120 mm) and thickness (0.2, 0.4 and 1 mm).

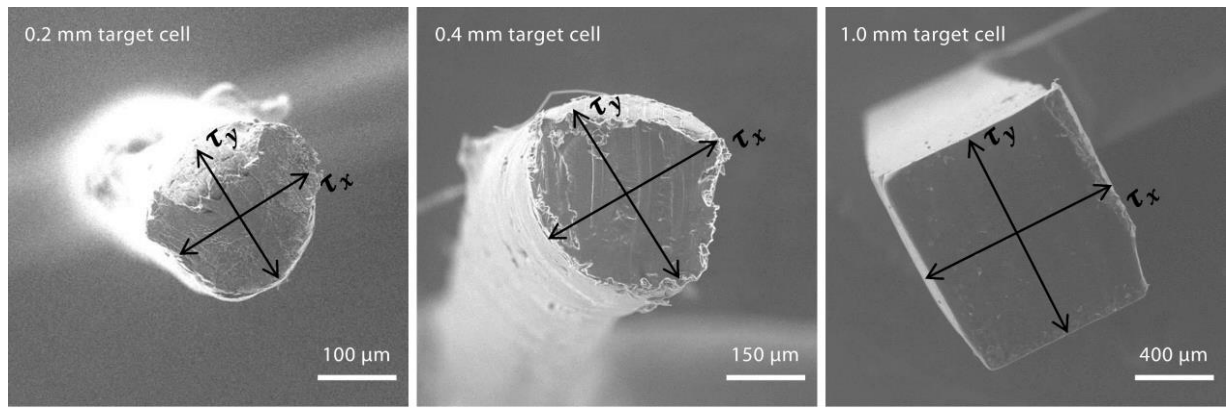

**Figure S7.** Microscopic images of target cells cross-section. Samples of three fabricated target cells with uniform cross-sectional thickness ($\tau_x \approx \tau_y$) of $\tau$ = [0.2, 0.4 and 1.0] mm.

**Section S4. Statistical analysis of the geometric study of the target cells**

To find the significance of each geometric parameter (of target cells) on entangled picking, we run a multi-dimensional statistical analysis called dominance analysis [31]. The dominance analysis creates a group of subset models in which it measures the contribution of each predictor (in our case the geometric parameters) and their relative importance which defines their statistical significance in determining the output (in our case the picked amount).

We use a Python library called "dominance-analysis" [43] to generate 8 subset models ($= 2^n - 1$, where $n = 3$ is the number of predictors in our case) including 3 models with only one predictor, 4 models with two predictors and 1 model with three predictors. We then compute the $R^2$ of each subset model and the complete model using McFadden's Pseudo-R Squared [44] - which compares the log likelihood of the full model ($L_{full}$) to the log likelihood of the model with only the intercept ($L_{null}$).

$$R^2 = 1 - \frac{\log(L_{full})}{\log(L_{null})}$$

The dominance analysis then measures the average additional contributions of each predictor by their incremental impact in the presence of all other predictors [43]. This is obtained by subtracting the R-squared value of a model with all other parameters from the R-squared value of the complete model and it is defined as "interactional dominance" (in our case the interactional dominance is as same as the total dominance for each predictor).

To conclude, we can measure the relative importance for each geometric parameter using their interactional dominance which is derived from the ratio of the interactional dominance in respect to the R-squared of the complete model:

$$\text{Relative Importance} = \frac{\text{Interactional Dominance}}{R^2 \text{ of complete model}}$$

In our analysis, we find the R-Squared of the complete model to equal to 0.794 and we find the percentage of relative importance for each geometric parameter (reported in Table S3).

**Table S3.** Dominance Analysis of the geometric parameters in entangled picking.

| Geometric parameters | Length | Thickness | Spike |
|---|---|---|---|
| Interactional Dominance | 0.591 | 0.130 | 0.072 |
| Percentage of Relative Importance | 74.41% | 16.45% | 9.14% |

**Section S5. Picking silicon rubber tubes with three core types**

In general, thickness and bending stiffness in objects have a proportional relationship, which is also the case for our target cells (refer to Table S1). Unfortunately, the geometric experiment does not allow us to decouple the effect of thickness from bending stiffness and understand how this parameter solely affects the ability of the targets to get entangled. Therefore, in an attempt to distinguish the entanglement caused by the material property from that caused by the geometric features, we set up a supplementary experiment. In particular, we control for geometry and frictional interactions by fabricating another set of target cells. These consist of geometrically identical silicon rubber tubes (with the same surface friction) which we fill with three different types of cores: i) nylon, ii) lead and iii) aluminium (see Fig. S8(A)). The core type gives each target a different bending stiffness (i.e., low, medium and high, respectively), despite having the same geometry and surface friction.

We use 0.8-mm-diameter strings of nylon, lead and aluminium to fit inside the silicon rubber tube as its core (with 1.5 mm outer diameter, 1 mm inner diameter, see Fig. S8(B)). Each core type represents a level of stiffness from lowest (nylon) with 3 GPa, to medium (lead) with 14 GPa and highest (aluminium) with 69 GPa modulus of elasticity [42].

The bending stiffness (EI) of a beam is given by the product of the modulus of elasticity (E) and the area moment of inertia (I) of its cross-section. The three tubes have the same area moment of inertia due to having identical geometrical parameters and only differ in terms of elasticity (we neglect the elasticity of the silicon rubber tube of 0.06 GPa).

The results reported in Fig. S8(C) show the picking mean and standard deviation of nylon, lead and aluminium target cells (picked using our approach, see Fig. S8(D)). The statistical analysis shows no significant change in variance of picking between these target cells and no significant change in the mean picked value between lead and aluminium target cells (refer to Table S4, here we consider statistical significance for a p-value below 0.01). Indicating that bending stiffness only affects the mean picked amount for targets with low elasticity. Therefore, due to the minor effect of bending stiffness on the overall picking performance, we assume that geometrical parameters are the main influencers to determine entanglement.

**Table S4.** Statistical analysis of the significance ($p < 0.01$) of mean and variance in picking target cells with different bending stiffnesses.

| Paired two target cells | T-test for mean | F-test for variance |
|---|---|---|
| Nylon (3 GPa) – Lead (14 GPa) | $p < 0.01$ | $p = 0.398$ |
| Lead (14 GPa) – Aluminium (69 GPa) | $p = 0.014$ | $p = 0.071$ |
| Nylon (3 GPa) – Aluminium (69 GPa) | $p < 0.01$ | $p = 0.111$ |

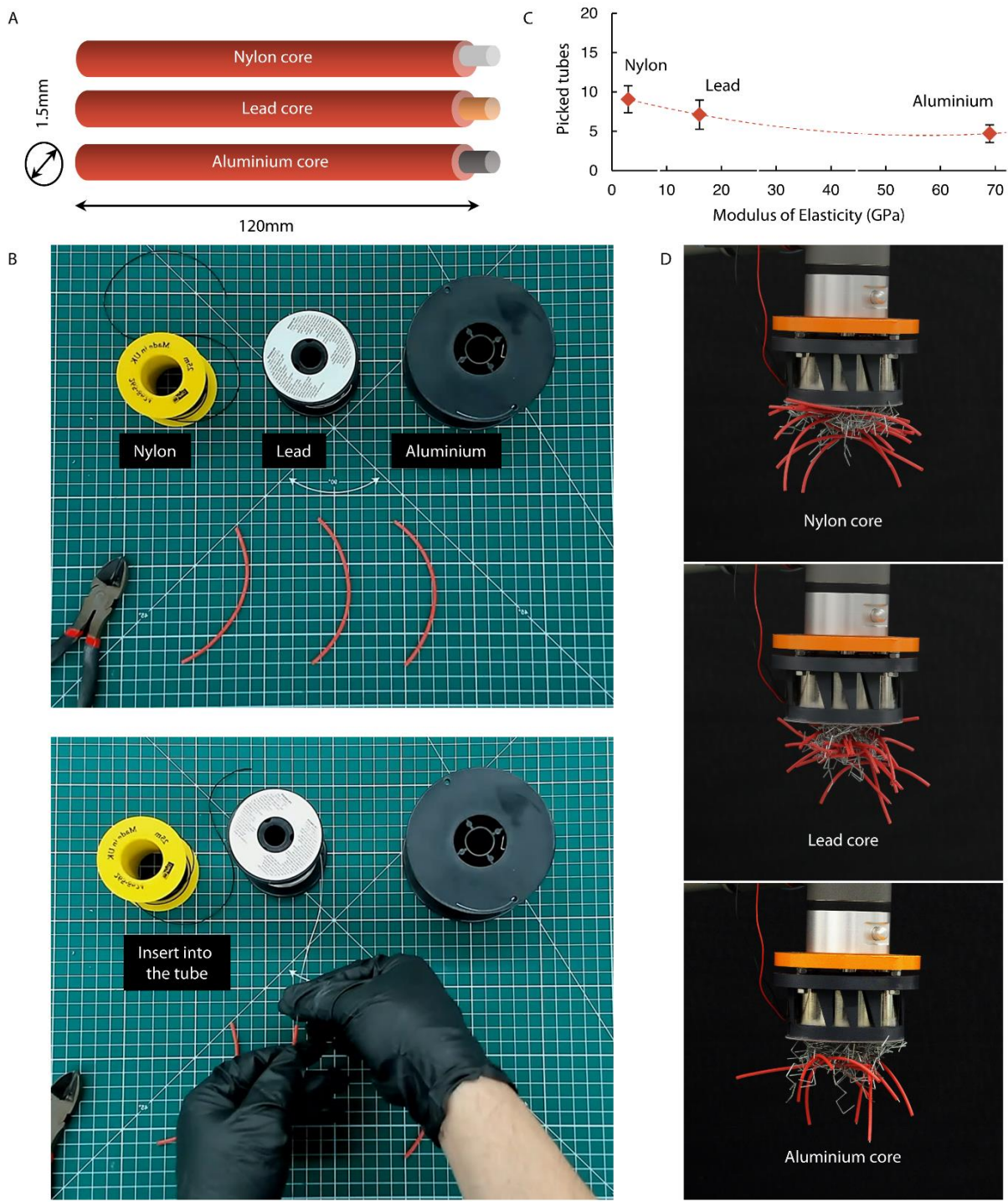

**Figure S8.** Fabrication and testing of entangled tubes. (**A**) Illustration of the fabricated Silicon rubber tubes with three core types. (**B**) The fabrication of three types of silicon tubes with three cores of Nylon, Lead and Aluminium. (**C**) The picked number of entangled tubes with respect to their modulus of elasticity. (**D**) Real-life images of picking three core-type silicon tubes using our approach.

**Section S6. Global probabilistic model to predict entangled picking**

Our linear regression model obtains the predicted picked amount ($\hat{y}$) from target cell's length ($\lambda$) and its thickness ($\tau$) in form of a logistic function, mathematically expressed as:

$$\hat{y} = \omega_1 \phi(\tau)\lambda + \omega_2$$

Where $\hat{y}$ is the predicted picked amount. $\omega_1$ and $\omega_2$ are the linear regression weights and $\phi(\tau)$ is obtained from:

$$\phi(\tau) = \frac{L}{1 + e^{-\theta_1(\tau - \theta_2)}} + L_0$$

Where $L$ is the curve's maximum value and $L_0$ is the offset or the minimum value. $\theta_1$ and $\theta_2$ are define curve's steepness and curve's sigmoidal midpoint, respectively.

We split our dataset into two sets of training (90% - $N_t$) and testing (10% - $N_T$). We fix the two parameters of $L$ and $L_0$ to 0.6 and 0.1 as the maximum and minimum values of the logistic curve, respectively. We optimise the other four parameters $\{\omega_1, \omega_1, \theta_1 \text{ and } \theta_2\}$ by minimising the normalised mean squared error (normalised by the standard deviation ($\sigma$)) using cross validation (leave-one-out) across the training data:

$$MSE_t = \frac{1}{N_t}\sum_{i=1}^{N_t}(y_i - \hat{y}_i)^2, \qquad NMSE_t = \frac{MSE_t}{\sigma}$$

Then, we verify the accuracy of our model by measuring the normalised mean squared error on the testing data (unseen by the model):

$$MSE_T = \frac{1}{N_T}\sum_{i=1}^{N_T}(y_i - \hat{y}_i)^2, \qquad NMSE_T = \frac{MSE_T}{\sigma}$$

We create two models for non-spiky (0-spike, $\sigma = 27.44$) and spiky (1-spike and 2-spikes, $\sigma = 38.27$) targets. We optimise both models and report their normalised mean squared error on training and testing dataset (refer to Table S5).

**Table S5.** The normalised mean squared error for the prediction models.

| Model | $NMSE_t$ (training) | $NMSE_T$ (testing – unseen by model) |
|---|---|---|
| non-spiky targets | 0.098 | 0.113 |
| spiky targets | 0.086 | 0.101 |

To generate a global probabilistic prediction model which can approximately predict a range of picking using our grasping approach, we particularly use Bayesian linear regression, which produces mean and uncertainty of prediction. We assume a gaussian distribution as prior for the weights of the linear regression ($\omega_1, \omega_1$). The Bayesian linear regression produces the mean (expected value of posterior distribution) and standard deviation (expected value of square root of variance) of the predictive distribution. We consider this standard deviation as the confidence interval of the model's prediction (66.66% confidence interval); for any potential targets with geometric parameters even beyond our target cells – which were used to train the model (see Fig. S9).

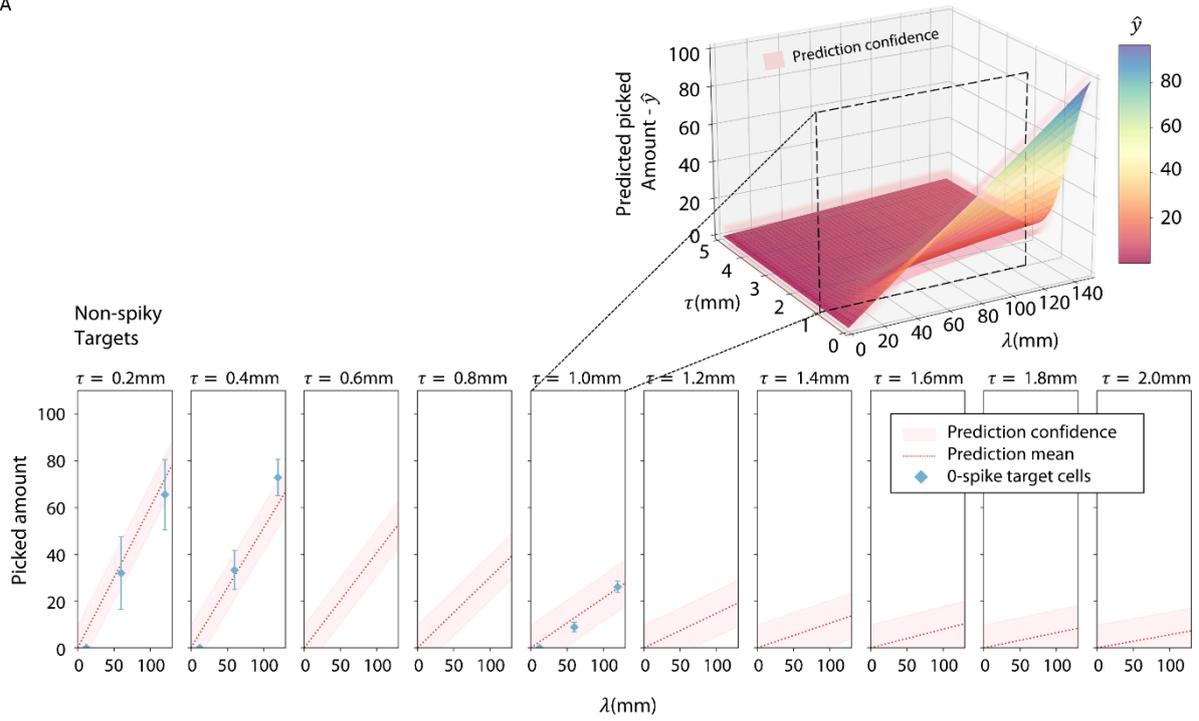

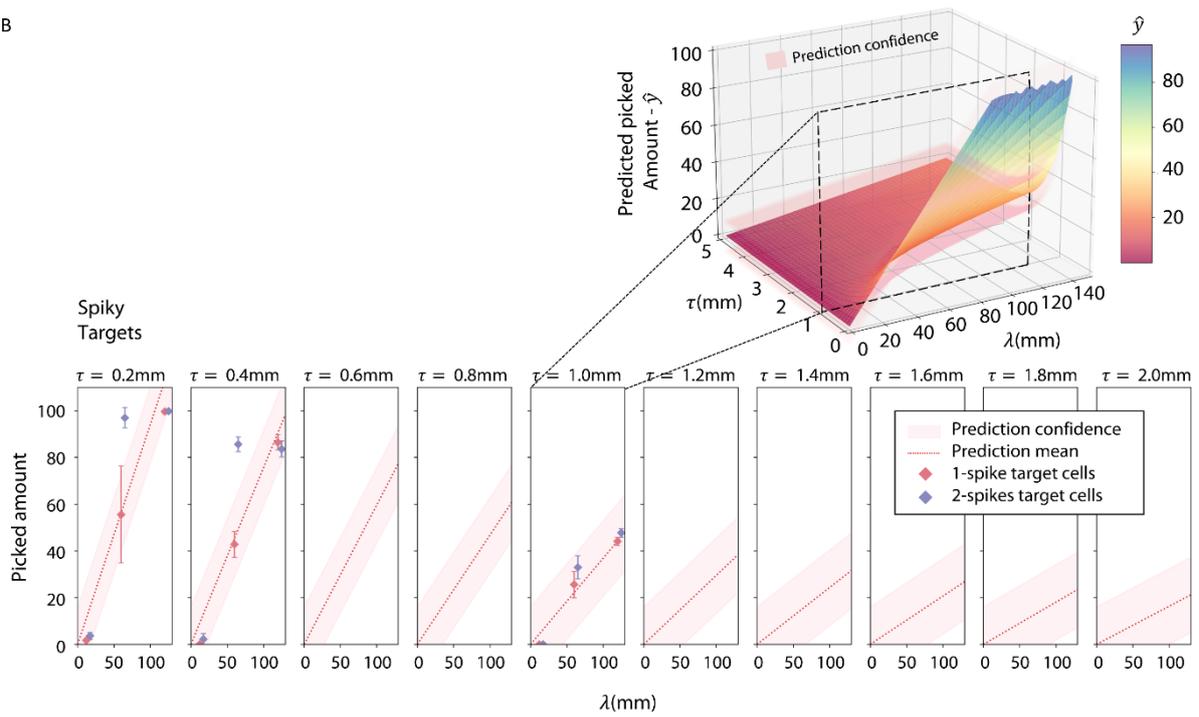

**Figure S9.** The global probabilistic prediction model for entangled picking. Including the confidence of prediction (standard deviation). **(A)** The 3D plot of predicted picked amount (top) and its 2D snapshots (bottom) for non-spiky targets. **(B)** The 3D plot of predicted picked amount (top) and its 2D snapshots (bottom) for spiky targets.

**Section S7. Grasping real-life targets with a robotic parallel gripper**

We use the same experimental setup and follow the same experimental procedure used for our grasping approach to pick real life targets with a parallel gripper (see Fig. S10). This includes using the same hemispherical container (plastic bowl) and having 100 units of obtainable targets inside.

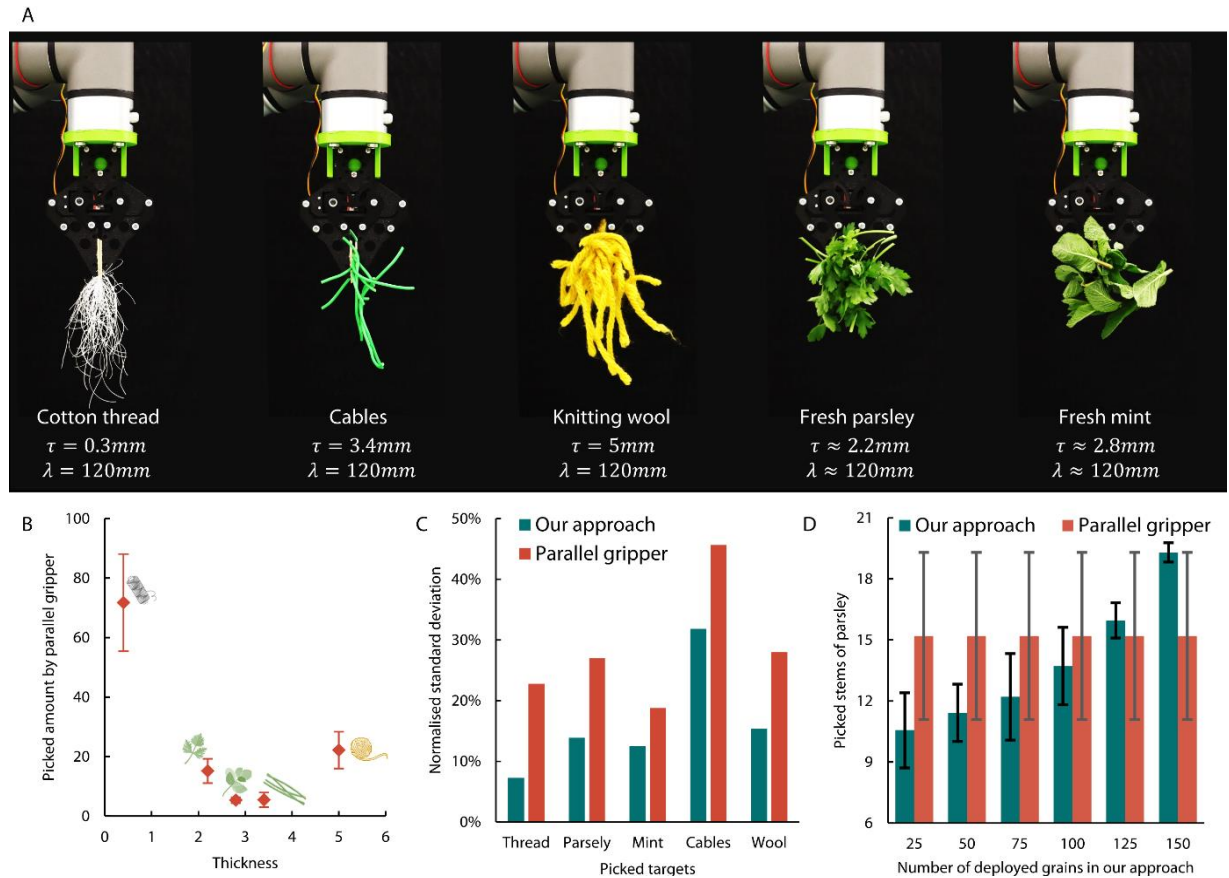

**Figure S10.** Picking real-life targets with a parallel gripper. **(A)** The real-life images of the picking of various targets with a parallel gripper. **(B)** The picking mean and standard deviation of 5 real-life targets with various thicknesses. **(C)** A comparison between our gripping approach and a parallel gripper in terms of their picking standard deviation (normalised by dividing the picked standard deviation by the picked mean for each target type – our approach features a lower standard deviation of 96% on average). **(D)** A comparison of mean and standard deviation of picking fresh parsley between the parallel gripper and our grasping approach with different number of deployed grains (from 25 to 150 grains) – reaching up to 11 times lower standard deviation in picking when 150 grains are deployed.

## Section S8. Fabrication of decomposable grains

Here we demonstrate the step-by-step fabrication method of two types of decomposable grains: a) ferromagnetic dismantlable grains (see Fig. S11) and b) ferromagnetic ice grains (see Fig. S12).

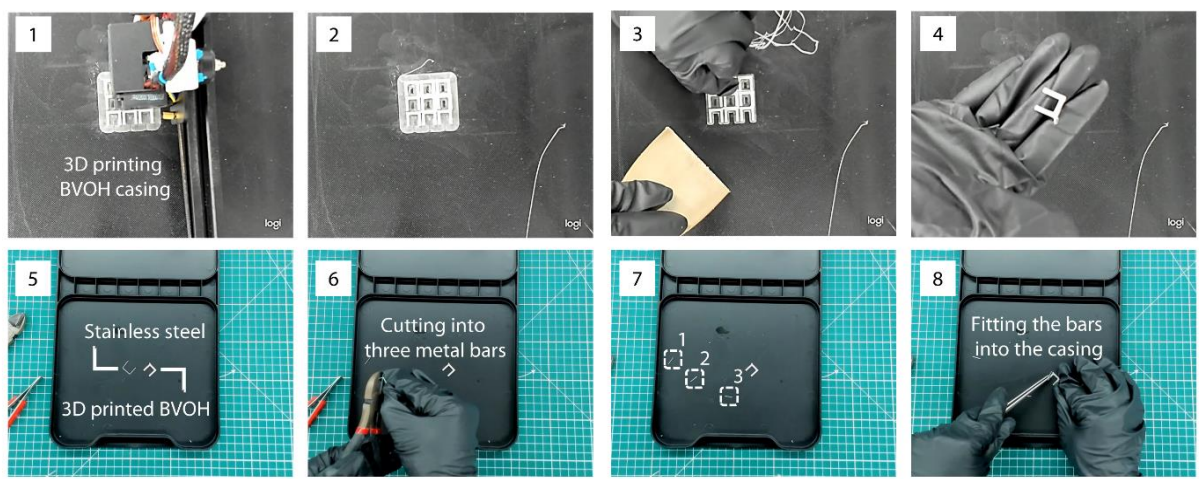

**Figure S11.** The fabrication process of ferromagnetic dismantlable grains. The grain casings are 3D printed using BVOH (Butenediol vinyl alcohol co-polymer, BASF) 3D printing filament (1-4). A stainless-steel staple is cut into three metal bars and tightly fitted into the BVOH casing (5-8).

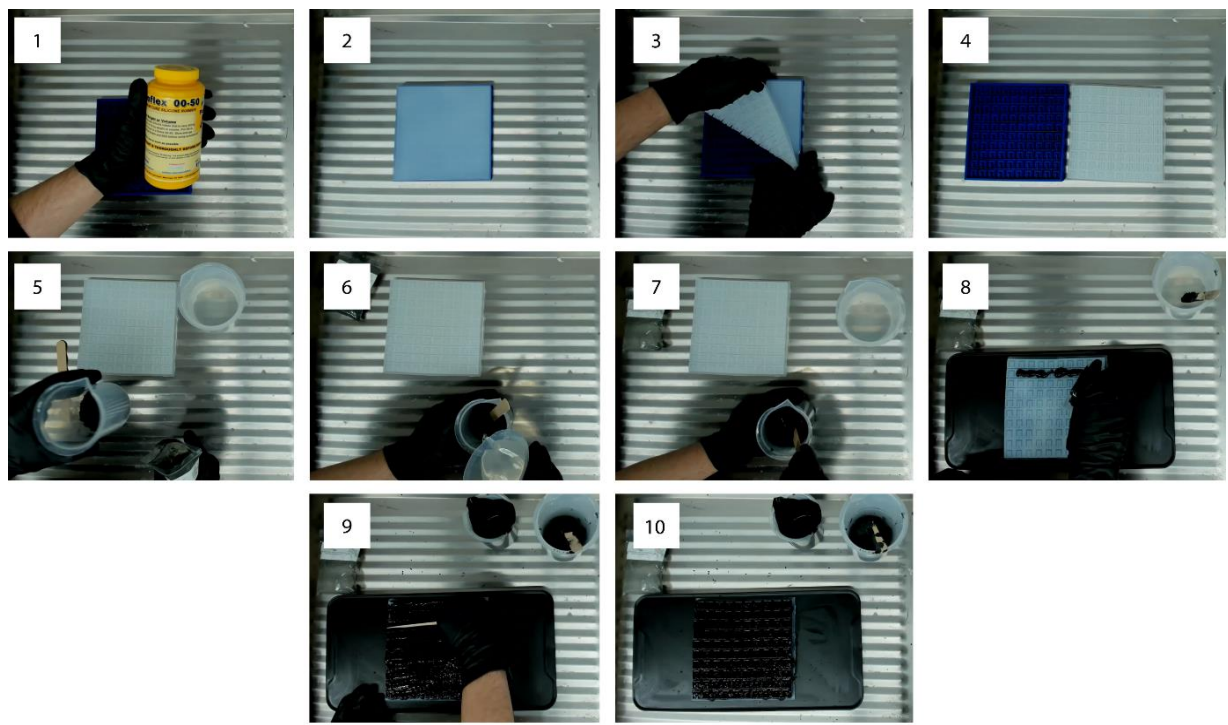

**Figure S12.** The fabrication process of ferromagnetic ice (FI) grains. We use Eco-flex 00-50 silicon elastomer (Smooth-On) for casting (1). The silicon is poured into a 3D printed moulding case (2). The newly fabricated silicon mould is removed from the 3D printed moulding case (3,4). A mixture of 66.6% water and 33.3% black iron oxide (ferromagnetic powder) is prepared (5-7). The mixture is poured onto the silicon mould and is spread evenly to completely fill the mould (8-10).

**Movie captions**

Movie S1 starts with illustrations from the design process of grains Each grain type is constructed from a 12 mm long and 1×1 mm thick segment. Then, extra segments are added to the initial segment to create spiky grains (1, 2 and 4 spikes). Moreover, two more families of grains are designed using the same process but with different orientation of spikes and introducing curvature in the initial segment. Overall, nine grain types are designed and are fabricated by laser cutting a 1 mm thick acrylic sheet.

Movie S2 demonstrates a video of three grain types (100 segments (4 g) of each) inside cylindrical tubes. The tubes are vertically removed to observe the integrity of the granular structure created by the grains. Two of the grain types' structures (0-spike and 4-spikes) collapse immediately while the structure created by 2-spike grains remains standing - demonstrating high structural integrity. Furthermore, we see the structural integrity of all nine grain types varying from the lowest with 33% to highest with 95% integrity.

Movie S3 shows a comparison between I-shaped and U-shaped grains when being picked by a robotic parallel gripper, in which the U-shaped grains are picked not only within the grasp of the parallel gripper but by tangling to each other. Additionally, the difference in picking for all nine grain types are demonstrated to visually observe the excessive picking which occurs due to tangling between some of the grain types.

Movie S4 displays the process of the geometric design (length, thickness and number of spikes) as well as the fabrication (laser cutting) of the target cells. Furthermore, the real-life images of the 27 fabricated target cells are demonstrated with a close-up look to visually observe the difference in deformation (stiffness) of cells with three different thicknesses.

Movie S5 demonstrates the experimental procedure of our grasping approach. It starts with picking ferromagnetic grains (made from stainless steel) by an electromagnet mounted to the end of the robotic arm. Then the grains are dropped onto non-ferromagnetic targets to interact. A quick proximity test is carried out to find the right vertical position with no gap between the electromagnet and the entangled cluster of grains and targets. Thereafter, the grains and a portion of entangled targets are picked with the electromagnet and moved to an empty bowl (dropping zone).

Movie S6 displays 5 continuous iterations of picking (using our grasping approach) real-life entangled targets (cotton thread, cables, knitting wool, fresh parsley, fresh mint and walnuts in a net bag).

Movie S7 demonstrates the fabrication process of two types of decomposable grains. The first type is created from 3D printed casing (water soluble) and three ferromagnetic metal bars (stainless-steel) fitted inside the casing. The second type is developed by mixing food safe ferromagnetic powder (black iron oxide) with water, then moulded into grain shapes in a fabricated silicon moulding case. Lastly, the moulded mixture is frozen to fully solidify.

Movie S8 shows picking of fresh parsley (using our grasping approach) with first type of decomposable grains (ferromagnetic dismantlable grains). The picked parsley and the grains are dropped into a bowl of water. The grains' casing reacts to water and starts dissolving, resulting in complete untangling of the grains from the fresh parsley.

Movie S9 demonstrates use of a permanent magnet (instead of an electromagnet) to pick fresh parsley with ferromagnetic ice grains. The picked pile is transferred to an empty bowl (dropping zone) and in less than 2 minutes, the water in the grains is almost fully melted, resulting in sudden drop of the picked fresh parsley with the ferromagnetic powder remaining on the magnet (minimising the contamination of fresh parsley).

**References**
(references [31], [39] and [40] can be found in the main article)